%% file: egpaper_for_review.tex
\documentclass[10pt,twocolumn,letterpaper]{article}
\usepackage{adjustbox}
\usepackage{multirow}
\usepackage{cvpr}
\usepackage{times}
\usepackage{epsfig}
\usepackage{graphicx}
\usepackage{amsmath}
\usepackage{amssymb}
\usepackage{graphicx}
\usepackage{stfloats}
\usepackage[table]{xcolor}
\usepackage{colortbl}
\usepackage{booktabs}
\definecolor{rowgray}{gray}{0.9}
\usepackage{float}
\usepackage{enumitem}
\usepackage[utf8]{inputenc}

\begin{document}

\title{PromptPrint: Behavioral Biometrics Through\\ Natural Language Prompting in LLMs}

\author{Shaiv Patel \quad Kartik Narayan \quad Vishal Patel\\
Johns Hopkins University, Baltimore, MD, USA\\
{\tt\small \{spate235, knaraya4, vpatel36\}@jh.edu}
}
\maketitle
\input{0_abstract}    
\input{1_Intro}
\input{2_Related_work}
\input{3_Dataset}
\input{3_Proposed_Work}

\input{5_Results}
\input{7_discussion}
\input{5_Conclusion}

{\small
\bibliographystyle{ieeetr}
\bibliography{egbib}
}

\end{document}

%% file: 0_abstract.tex
\begin{abstract}

Authorship attribution research has traditionally focused on long-form, expressive texts; however, interactions with large language models (LLMs) are typically brief and task-driven prompts. This raises a fundamental question: do such prompts contain a stable, author-identifiable, and distinctive signal?  We introduce \textbf{PromptPrint}, a systematic study of prompt-based identity, the hypothesis that a user’s habitual vocabulary, syntax, and discourse patterns form a learnable behavioral biometric. Using 20,680 real prompts from 1,034 users, we establish three key findings. First, lexical representations significantly outperform semantic encoders, supporting the ``lexical stability hypothesis": identity is primarily encoded in surface-level word choice rather than abstract intent. Second, stylometric features exhibit a ``uniqueness–consistency paradox": users are highly distinctive across the population, yet behaviorally inconsistent across contexts. Third, adversarial analysis reveals a clear vulnerability spectrum: identity signals are robust to minor lexical perturbations but degrade substantially under semantic paraphrasing. Overall, our results demonstrate strong identification performance at scale, establishing prompt-based identity as a viable behavioral biometric. This work introduces a new perspective on user modeling in LLM interactions, with important implications for security and privacy. Data and code will be released upon the acceptance of our work.

\end{abstract}

%% file: 1_Intro.tex
\section{Introduction}
\label{sec:intro}

Large language models (LLMs) have made natural language a central interface for everyday human-computer interaction through \emph{prompting}. Each prompt is a deliberate linguistic act in which a user chooses how to articulate an instruction to a machine. While authorship attribution research has established that writing inherently carries identity~\cite{stamatatos2009survey}, LLM prompts impose extreme linguistic compression and are primarily \emph{instrumental} in nature-spanning only one to three sentences, stripped of narrative elaboration, and strictly task-directed. Traditional authorship attribution assumes sufficient length and expressive intent; prompts routinely violate both. It remains unclear whether stylometric identity signals persist under these constraints, and which feature spaces best capture them.

We introduce \textbf{PromptPrint} (Figure~\ref{fig:promptprint}) and make four primary 
contributions:

\begin{enumerate}[noitemsep]
    \item We formalize \emph{prompt-based identity} as a novel soft biometric modality and establish the first evaluation protocol for this setting, grounded in standard biometric metrics (EER, $d'$, ROC-AUC).
    \item We conduct a controlled evaluation of lexical, semantic, and stylometric feature spaces for prompt identification across 1,034 real users using 5-fold cross-validation.
    \item We identify and characterize a \emph{uniqueness-consistency paradox} within stylometric features where high inter-user separability paradoxically coexists with low intra-user consistency and contextualize it through signal detection theory.
    \item We evaluate the stability of prompt as soft biometric against three adversarial attacks of escalating severity: synonym substitution, homoglyph substitution, and full semantic paraphrase. We establish that full rewriting severely degrades ensemble identification (Top-1=$0.429$, EER=$0.703$), whereas synonym substitution causes negligible impact ($\Delta$Top-1=$-$0.001). These results indicate that prompt-based identity signals are concentrated at the token-surface level and that semantic paraphrasing poses a substantially stronger challenge than minor lexical perturbations.
\end{enumerate}

Finally, because prompt-based identity can operate passively and without explicit user awareness, it raises substantial privacy concerns. We frame this work as identifying a previously uncharacterized surveillance surface and providing a foundation for future defensive research.

%% file: 2_Related_work.tex
\begin{figure*}[t]
  \centering
  \includegraphics[width=1.7\columnwidth]{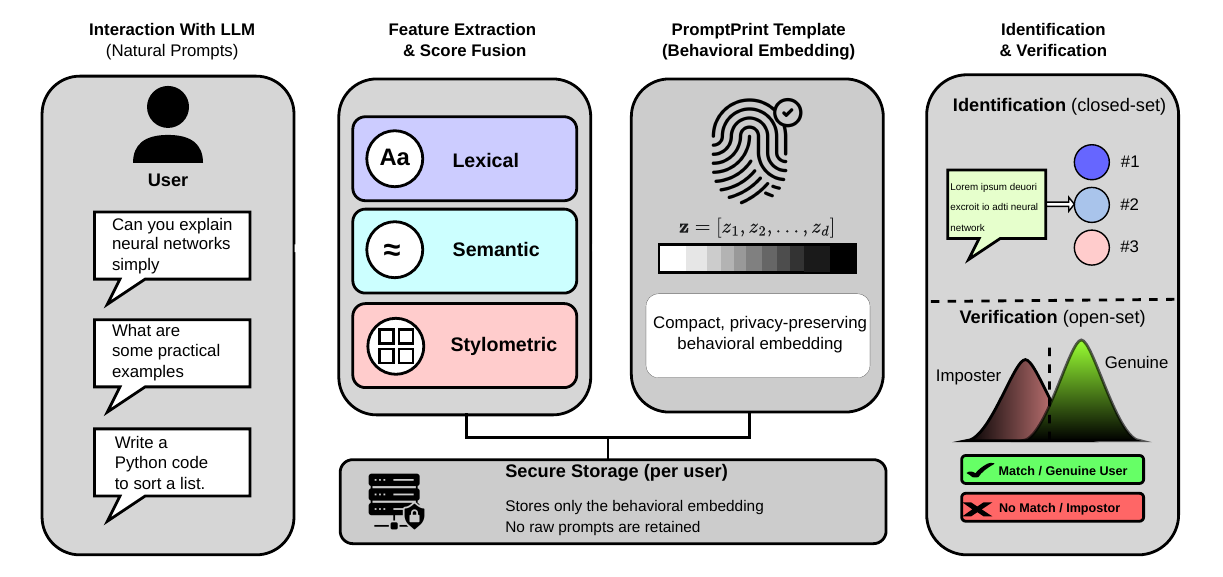}
  \caption{LLM prompts as a soft biometric for authentication overview.  A user's prompt behavior can capture how they think, ask, and express -- forming a stable behavioral signature that can be used for authentication.  }
  \label{fig:promptprint}
\end{figure*}

\section{Related Work}
\label{sec:related}

\noindent{\bf{Authorship Attribution.}} Traditional authorship attribution relies on feature spaces ranging from character n-grams~\cite{stamatatos2009survey} and function word frequencies~\cite{koppel2011authorship} to syntactic patterns~\cite{raghavan2010authorship} and deep representations~\cite{fabien2020bertaa}. A recent comprehensive survey ~\cite{huang2025authorship} reviews these methodologies in the context of LLM-generated text, while Rivera-Soto et al. ~\cite{rivera2021learning} proposed learning universal authorship representations via contrastive learning for cross-domain authorship verification. However, these methodologies are fundamentally predicated on analyzing expressive text such as essays, forum posts, or literary prose of sufficient length.\\
\noindent{\bf{Behavioral Biometrics.}} 
Behavioral biometrics traditionally isolates persistent identity signals in modalities such as keystroke dynamics~\cite{monrose1997authentication,roth2014continuous}, mouse movement~\cite{ahmed2007new}, touchscreen interaction~\cite{frank2012touchalytics}, and gait~\cite{boulgouris2006gait}. These modalities share a critical property with our framework: they operate \emph{passively}, requiring no explicit enrollment gesture from the user. We evaluate our findings using standard verification and identification protocols for behavioral biometrics. However, we extend the broader biometrics paradigm by introducing LLM prompts as a distinctly \emph{linguistic} behavioral channel, establishing a new modality that prior research has yet to explore.\\
\noindent{\bf{LLM Security and Privacy.}} 
Contemporary research in LLM security predominantly focuses on semantic content, analyzing vulnerabilities through adversarial attacks~\cite{zou2023universal}, prompt injection~\cite{perez2022ignore}, and membership inference~\cite{carlini2021extracting}. Concurrently, re-identification via writing style has been explored in the contexts of de-anonymization~\cite{narayanan2008robust} and browser fingerprinting~\cite{eckersley2010unique}. Moreover, Staab et al. ~\cite{staab2023beyond} demonstrated that LLMs can infer personal attributes from user text at scale, highlighting inference-based privacy risksbeyond training data memorization. Our framework bridges these domains: we isolate stable stylistic behavior in prompts as a persistent identity channel, operating independently of the semantic intent or payload of the query.\\
\noindent{\bf{Prompting as a Distinct Behavioral Signal.}} 
To our knowledge, no prior work investigates prompting behavior as a measurable behavioral biometric. The most adjacent line of research applies stylometric analysis to short social media texts for authorship verification ~\cite{tyo2022state}, ~\cite{wegmann2022same}. However, social media content remains fundamentally expressive and self-initiated. Prompts, conversely, are reactive, task-directed constraints.

%% file: 3_Dataset.tex
\section{Dataset}
\label{sec:dataset}

We utilize WildChat-1M~\cite{zhao2024wildchat}, a public corpus comprising over one million authentic ChatGPT conversations collected via a proxy interface, with users distinguished by SHA-256 hashed IP addresses (\texttt{hashed\_ip}). These interactions were captured in the wild without users’ awareness of subsequent analysis, so the dataset preserves genuine prompting habits and avoids the behavioral artificiality inherent to laboratory-constrained datasets. The corpus is publicly released under an open license.
 \subsection{Extraction Protocol}
To isolate the user’s baseline identity signal, we extract only the first user turn from each conversation: the initial, autonomous instruction provided prior to any LLM-guided follow-up. This methodological constraint is vital for assessing behavioral stability, as users in multi-turn dialogues frequently exhibit linguistic accommodation by subconsciously mirroring the model’s vocabulary and structural register. Such adaptation would contaminate the underlying stylometric fingerprint. We subsequently apply four quality filters to the isolated prompts:
\begin{itemize}[noitemsep]
    \item Language: English only (\text{language == `English'})
    \item \emph{Minimum length:} $\geq$10 characters
    \item \emph{Maximum length:} $\leq$2,000 characters
    \item \emph{Content:} Exclusion of URL-only and code-block-only prompts
\end{itemize}
Users with fewer than 20 qualifying prompts are excluded. Retained users are ranked by total prompt count, and the top 1,500 are evaluated; exactly 1,034 users meet the strict 20-prompt threshold with complete qualification.

\subsection{Statistics}
\begin{figure}[htp!]
  \centering
  \includegraphics[width=1\columnwidth]{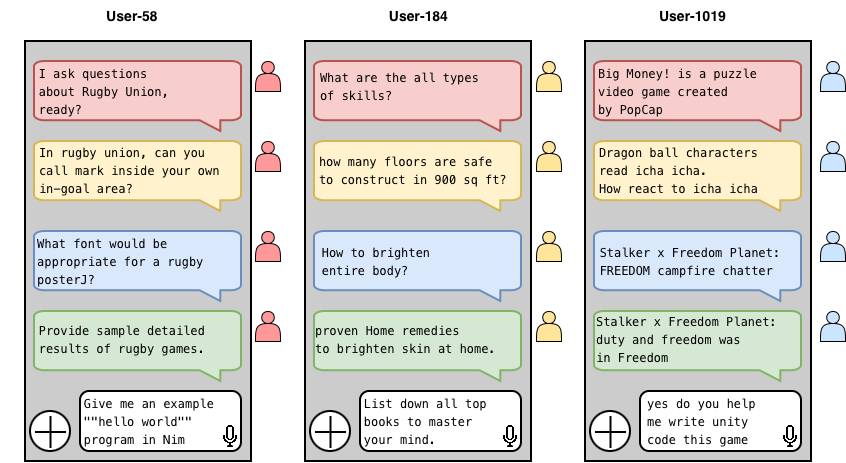}
  \caption{Samples prompts from the WildChat-1M dataset.}
  \label{fig:samplesPrompts}
\end{figure}

The final evaluation dataset comprises 20,680 prompts distributed uniformly across 1,034 users, with 20 prompts per user. The prompt length distribution: mean 187 characters, standard deviation 312, and median 98 characters, quantitatively reflects the highly compressed, task-directed nature of real-world prompting behavior. Sample prompts from the WildChat-1M dataset are shown in Figure~\ref{fig:samplesPrompts}.

\subsection{Ethical Compliance}
\label{sec:ethics_data}
WildChat-1M is explicitly released for research applications. All user identities are cryptographically hashed at the source, ensuring no personally identifiable information (PII) is retained or processed. As this study involves retrospective analysis of anonymized public data without direct human subject interaction, it is exempt from institutional review board (IRB) approval under standard guidelines. We note that WildChat-1M's hashed IP-based user proxies are a dataset-level constraint, not a methodological choice. Label noise from shared and dynamic IPs makes identification strictly harder, so our metrics represent a conservative lower bound on true performance; datasets with verified user identities would 
yield stronger signal detection. We address the broader ethical implications of establishing prompt as a soft biometric channel 
in Section~\ref{sec:discussion}.

%% file: 3_Proposed_Work.tex
\section{PromptPrint: Modeling and Inference}
\label{sec:method}

The PromptPrint behavioral embedding $\mathbf{z} = [z_1, z_2, 
\ldots, z_d]$ denotes the feature vector extracted from a user's 
prompt under a given representation space (lexical, semantic, or 
stylometric), used to construct per-user prototype vectors for 
identification and verification.

\subsection{Features}
Given user prompts, we investigate three distinct feature 
representations to identify which signals reliably capture 
stable behavioral identity.

\noindent\textbf{Lexical (TF-IDF) Features.}
We extract bag-of-words representations using unigrams and bigrams, with sublinear term-frequency scaling ($\log(1+\mathrm{tf})$), where $\mathrm{tf}$ denotes the raw term frequency and the logarithmic transform attenuates the influence of high-frequency terms.  Features are constrained by a minimum document frequency of 2 and a vocabulary capped at 30,000 features~\cite{sparck1972statistical}. This space isolates the specific vocabulary and phrasal habits unique to a user's generative behavior.\\
\noindent\textbf{Semantic (SBERT) Features.}
We utilize 768-dimensional, L2-normalized sentence embeddings derived from \texttt{intfloat/e5-base}~\cite{wang2022text}. E5 is trained via contrastive learning on large-scale text pairs for general-purpose dense retrieval. This space encodes the overarching semantic intent of prompts, independent of specific surface phrasing. We select \texttt{e5-base} as it demonstrated stable performance across preliminary encoder comparisons, and its computational footprint is compatible with 5-fold cross-validation at scale; larger encoders would likely strengthen the semantic baseline but would not alter the core finding that lexical representations fundamentally outperform semantic encoders for this task.\\
\noindent\textbf{Stylometric Features.}
Following established authorship attribution conventions~\cite{stamatatos2009survey}, we extract ten hand-crafted features: average word length, vocabulary richness (type-token ratio), punctuation ratio, stopword ratio, noun/verb/adjective ratios (via Penn Treebank POS tags using NLTK~\cite{bird2009natural}), average sentence length, uppercase ratio, and digit ratio. This quantifies surface-level writing patterns strictly independent of semantic content.

\subsection{Models and Classifiers}

We train six systems for comparative evaluation. All training and benchmarking were performed on NVIDIA A100 Tensor Core GPUs.

\vspace{2pt}
\noindent\textbf{(TF-IDF)+LR.} A TF-IDF vectorizer is paired with a multinomial Logistic Regression classifier using the L-BFGS solver, $C=1.0$, and balanced class weights.

\vspace{2pt}
\noindent\textbf{SBERT-NN.} A frozen SBERT encoder (768-d) feeding a three-layer MLP. The architecture comprises of an input projection (768$\to$384), a residual block (384$\to$384), compression layers (384$\to$192$\to$96), and a final classification head (96$\to$N). The encoder remains strictly frozen to prevent catastrophic forgetting~\cite{mccloskey1989catastrophic} on the constrained per-user training sets. Training utilizes label-smoothed cross-entropy ($\varepsilon$=0.1), the AdamW optimizer~\cite{loshchilov2017decoupled} ($\eta$=1e-4, weight decay=0.01), a cosine annealing learning rate schedule, and early stopping (patience=10).

\vspace{2pt}
\noindent\textbf{Stylo-NN.} This utilizes an identical MLP architecture, with the input projection adapted to $10\to384$ to process the 10-dimensional stylometric feature vector, exclusively isolating surface-level structural habits.

\vspace{2pt}
\noindent\textbf{Combined-NN.} SBERT and stylometric features are concatenated (778-d) prior to feeding the standard MLP, directly testing feature-level fusion capabilities.

\vspace{2pt}
\noindent\textbf{Ensemble.} A score-level fusion mechanism defined as:
\setlength{\belowdisplayskip}{0pt} \setlength{\belowdisplayshortskip}{0pt}
\setlength{\abovedisplayskip}{0pt} \setlength{\abovedisplayshortskip}{0pt}
\begin{equation}
  P_{\text{ens}} = \alpha \cdot P_{\text{SBERT}} + (1-\alpha) \cdot P_{\text{TF-IDF}}
  \label{eq:ensemble}
\end{equation}
where $P_{\text{SBERT}}$ and $P_{\text{TF-IDF}}$ are the class probability distributions produced by each component, $P_{\text{ens}}$ is the fused output distribution, and $\alpha \in \{0.0, 0.1, \ldots, 1.0\}$ is the interpolation weight swept to identify the optimal fusion ratio.

\vspace{2pt}
\noindent\textbf{CharNgram+SVM.} Character $n$-gram features (2–5)~\cite{stamatatos2009survey} coupled with LinearSVC~\cite{fan2008liblinear}, serving as the standard authorship attribution baseline.

\subsection{Evaluation Protocol}
\noindent\textbf{Data splits.} We employ 5-fold cross-validation. In each fold, the population is partitioned into seen users (80\%, $\approx$828 users) and unseen users (20\%, $\approx$206 users). Seen users contribute 16 training, 2 validation, and 2 test prompts each. Unseen users constitute the strict impostor set for the verification evaluation. All splits are computationally validated to guarantee zero user overlap between seen and unseen cohorts.

\vspace{2pt}
\noindent\textbf{Identification metrics.}
We evaluate closed-set Top-k accuracy over seen users 
reporting Top-1, Top-3, and Top-5 accuracy. Closed-set evaluation assumes that the query user necessarily exists within the gallery; therefore, we also address open-set identification, where the system must determine whether a query belongs to the gallery at all, using the verification metrics detailed below.

\vspace{2pt}
\noindent\textbf{Verification metrics.}
Genuine scores are calculated as the cosine similarities between test embeddings and per-user prototype vectors (the mean train embedding). Impostor scores represent the maximum cosine similarities of unseen-user embeddings against the gallery. Open-set performance is quantified using the Equal Error Rate (EER), defined as the threshold at which the false-accept rate equals the false-reject rate. We report ROC-AUC, EER, and the sensitivity index $d'$:
\begin{equation}
d' = \frac{|\mu_g - \mu_i|}{\sqrt{\tfrac{1}{2}(\sigma_g^2 + \sigma_i^2)}},
\label{eq:dprime}
\end{equation}
where $\mu_g, \sigma_g$ denote the mean and standard deviation of genuine scores, and $\mu_i, \sigma_i$ denote those of the impostor scores. Verification metrics are computed uniformly across all methods via cosine similarity between test embeddings and per-user prototype vectors (mean train embeddings), with EER and AUC derived from the resulting genuine and impostor score distributions. For SBERT-based models, raw $768-d$ encoder embeddings are used directly; for (TF-IDF)+LR, the sparse TF-IDF vectors serve as the embedding space. Identification 
metrics utilize the final classification layer predictions across all methods.

\subsection{Adversarial Protocol}
To assess the biometric stability of prompt-based identity, we evaluate three threat models. The first two attacks are applied at varying perturbation rates $r \in \{5\%, 10\%, 20\%, 30\%\}$, while the third executes a complete rewrite:\\
\noindent\textbf{Homoglyph attack.}
Latin characters are replaced with visually identical Unicode homoglyphs (e.g., Latin `a' $\to$ Cyrillic `a', Latin `e' $\to$ Cyrillic `e'). This mechanism disrupts token recognition without altering the visual appearance of the text, forcing the tokenizer to produce out-of-vocabulary or fragmented subword units for targeted tokens.\\
\noindent\textbf{Synonym attack.}
Adjectives and adverbs are substituted with WordNet~\cite{fellbaum1998wordnet} synonyms (selected randomly from first-sense synset lemmas). This attack preserves the token-level surface form while scrambling modifier-level choices. Content words (nouns, verbs) and structural function words are strictly preserved to isolate and test the modifier-level identity signal.\\
\noindent\textbf{Paraphrase attack.}
The entire prompt is regenerated utilizing a neural paraphrase model
(\texttt{tuner007/pegasus\_paraphrase}~\cite{zhang2020pegasus}). This preserves the core semantic intent while aggressively over-writing the user's surface vocabulary, phrasal structures, and idiosyncratic habits. As a full-surface attack, no original token is guaranteed to survive. This is applied uniformly (rate=100\%) to benchmark the channel's vulnerability to maximum-effort surface erasure.

%% file: 5_Results.tex
\definecolor{rowgray}{gray}{0.9}

\section{Experimental Results}
\label{sec:results}

\subsection{Main Results}

Table~\ref{tab:main_results} reports 5-fold cross-validation results across all methods. All approaches substantially exceed the random 
baseline (Top-1=0.097\% for 1,034 users)

\begin{table*}[t]
\centering
\caption{%
  \textbf{Main Results.} 5-fold cross-validation results corresponding to 1,034 users.
  Best identification result in \textbf{bold}; best verification result in \textit{italic}.
  $\uparrow$ higher is better, $\downarrow$ lower is better.
  Random chance Top-1 $= 0.097\%$.
  TAR derived via ROC interpolation per ISO/IEC~19795.
}
\label{tab:main_results}
\adjustbox{max width=\textwidth}{%
\begin{tabular}{|l|c|c|c|c|c|c|c|c|c|}
\hline
\multirow{2}{*}{\textbf{Method}}
  & \multicolumn{3}{c|}{\textbf{Top-$k$ Accuracy $\uparrow$}}
  & \multirow{2}{*}{\textbf{AUC $\uparrow$}}
  & \multirow{2}{*}{\textbf{EER $\downarrow$}}
  & \multirow{2}{*}{\textbf{d$'$ $\uparrow$}}
  & \multicolumn{3}{c|}{\textbf{TAR @  FAR $\uparrow$}} \\
\cline{2-4}\cline{8-10}
  & \textbf{Top-1} & \textbf{Top-3} & \textbf{Top-5}
  & & &
  & \textbf{@1\%} & \textbf{@0.1\%} & \textbf{@0.01\%} \\
\hline
\multicolumn{10}{|l|}{\cellcolor{rowgray}\textit{Proposed}} \\
\hline
Ensemble ($\alpha$=0.7)
  & \textbf{0.642$\pm$0.004} & \textbf{0.755$\pm$0.004} & \textbf{0.793$\pm$0.005}
  & 0.603$\pm$0.005 & 0.421$\pm$0.005 & 0.471$\pm$0.009
  & 0.161$\pm$0.027 & 0.061$\pm$0.043 & 0.030$\pm$0.029 \\
\hline
Combined-NN
  & 0.518$\pm$0.008 & 0.621$\pm$0.004 & 0.664$\pm$0.004
  & 0.537$\pm$0.007 & 0.460$\pm$0.004 & 0.097$\pm$0.023
  & 0.172$\pm$0.027 & 0.070$\pm$0.042 & 0.056$\pm$0.032 \\
\hline
SBERT-NN
  & 0.524$\pm$0.009 & 0.612$\pm$0.003 & 0.650$\pm$0.006
  & 0.536$\pm$0.003 & 0.464$\pm$0.002 & 0.159$\pm$0.012
  & \textbf{0.203$\pm$0.044} & \textit{0.061$\pm$0.039} & 0.028$\pm$0.021 \\
\hline
Stylo-NN
  & 0.249$\pm$0.009 & 0.337$\pm$0.011 & 0.383$\pm$0.009
  & 0.276$\pm$0.006 & 0.684$\pm$0.008 & \textbf{0.950$\pm$0.022}
  & 0.083$\pm$0.007 & 0.045$\pm$0.015 & \textit{0.039$\pm$0.017} \\
\hline
\multicolumn{10}{|l|}{\cellcolor{rowgray}\textit{Baselines}} \\
\hline
(TF-IDF)+LR
  & 0.630$\pm$0.003 & 0.733$\pm$0.003 & 0.766$\pm$0.006
  & \textit{0.623$\pm$0.005} & \textit{0.381$\pm$0.006} & \textit{0.671$\pm$0.023}
  & \textit{0.166$\pm$0.043} & 0.034$\pm$0.009 & 0.023$\pm$0.016 \\
\hline
CharNgram+SVM
  & 0.542$\pm$0.006 & 0.632$\pm$0.007 & 0.660$\pm$0.007
  & 0.568$\pm$0.008 & 0.452$\pm$0.008 & 0.259$\pm$0.031
  & 0.147$\pm$0.024 & 0.057$\pm$0.046 & 0.022$\pm$0.016 \\
\hline
\end{tabular}%
}
\end{table*}

\vspace{4pt}
\noindent\textbf{Finding 1 - Lexical properties dominate semantic intent.}
(TF-IDF)+LR outperforms SBERT-NN across all verification metrics despite utilizing no contextual reasoning ($d'$=0.671 vs. 0.159, EER=0.381 vs. 0.463). This robustly supports the \emph{Lexical Stability Hypothesis}: sentence encoders optimized for semantic similarity~\cite{wang2022text} actively collapse semantically equivalent phrasings thereby destroying the identity signal while TF-IDF preserves the specific, idiosyncratic token choices that define a user's habitual behavior. This finding aligns with Wegmann et al. ~\cite{wegmann2022same}, who demonstrated that standard text embeddings conflate topical similarity with authorial style, and with Tyo et al. ~\cite{tyo2022state}, who independently found that traditional n-gram models outperform neural approaches on the majority of authorship attribution benchmarks. Paraphrase invariance is an intentional inductive bias for semantic search, but it proves fundamentally misaligned for biometric identification.

Notably, while the ensemble achieves the highest identification accuracy (Top-1=64.2\%), (TF-IDF)+LR retains the best verification performance (EER=0.381 vs.\ 0.421, $p{<}0.001$, paired $t$-test), confirming that $\alpha{=}0.7$ optimises for rank-based retrieval rather than verification threshold calibration; this trade-off is analyzed in Section~\ref{sec:alpha}.

\vspace{4pt}
\noindent\textbf{Finding 2 - Instrumental text requires a distinct biometric paradigm.}
While CharNgram+SVM serves as the standard short-text baseline for authorship attribution~\cite{stamatatos2009survey}, our prompt-specific ensemble demonstrates a clear behavioral divergence, identifying users with 64.2\% accuracy compared to the baseline's 54.2\%. This 10-point delta empirically confirms that the highly compressed, task-directed nature of prompting cannot be fully captured by legacy expressive-text models; it requires feature spaces explicitly tuned to instrumental linguistic habits.

\vspace{4pt}
\noindent\textbf{Finding 3 - Feature-level fusion degrades component efficacy.}
Combined-NN ($d'=0.096$) yields weaker separation than both SBERT-NN ($d'=0.159$) and Stylo-NN ($d'=0.950$) in isolation. The ten stylometric features comprise merely 1.3\% of the 778-dimensional joint vector; therefore, the MLP classifier lacks a structural mechanism to up-weight this low-dimensional signal against the high-variance 768-dimensional SBERT embeddings. Score-level fusion through the Ensemble model effectively circumvents this bottleneck by preserving the independent discriminative power of each modality.

\begin{figure*}[b]
  \centering
  \includegraphics[width=0.32\linewidth]{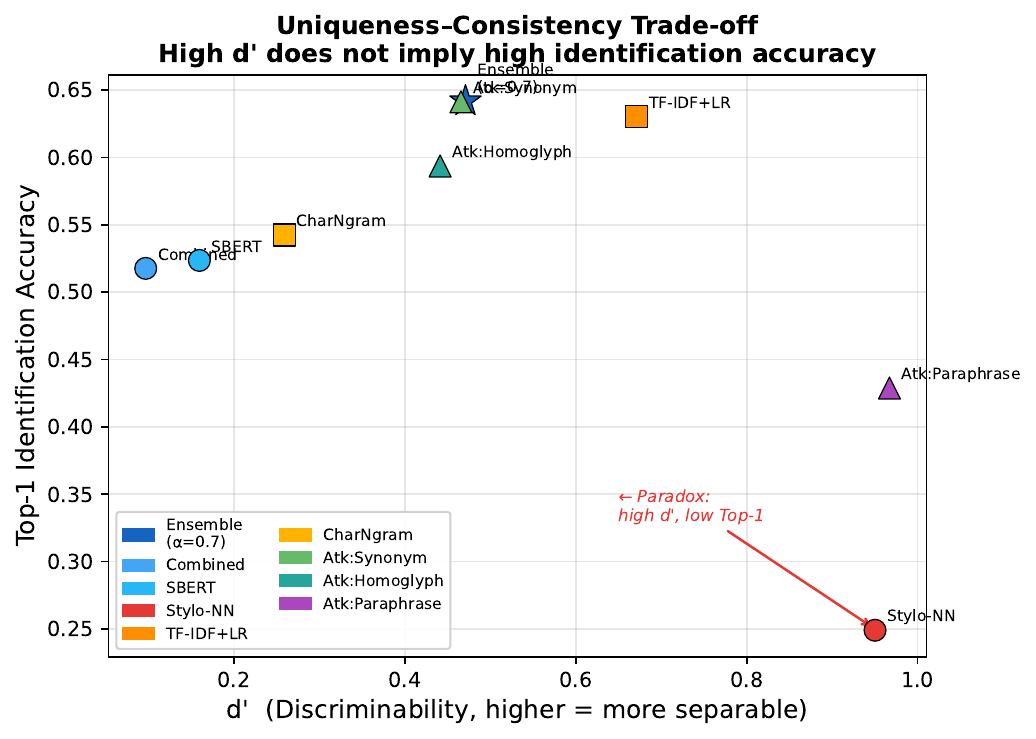}
  \hfill
  \includegraphics[width=0.32\linewidth]{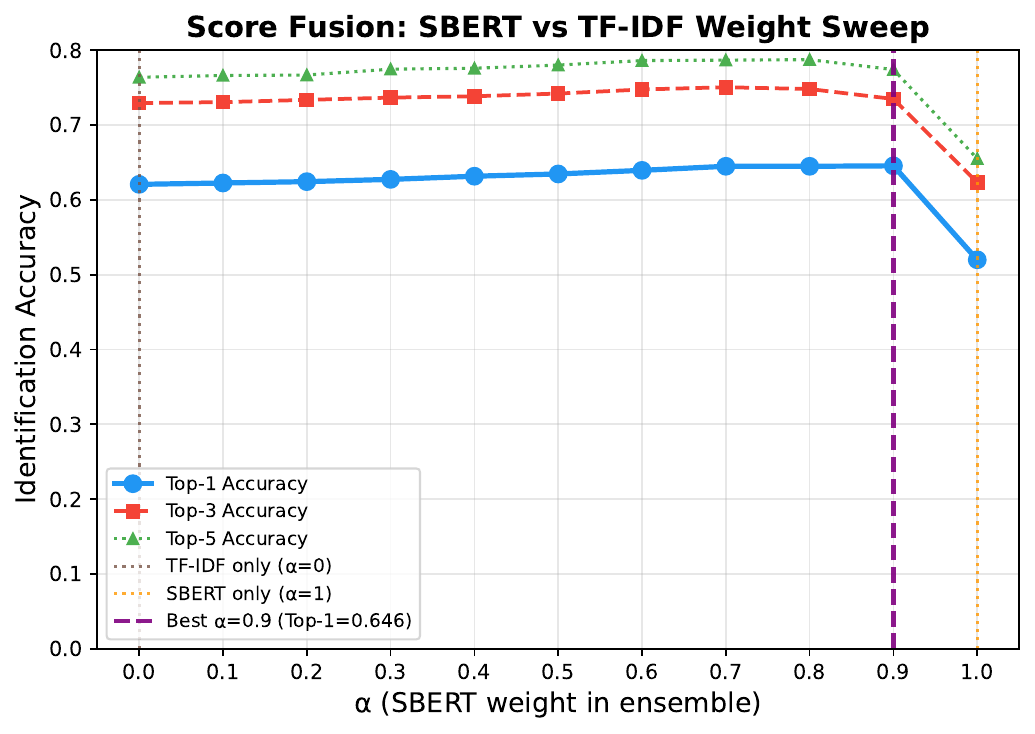}
  \hfill
  \includegraphics[width=0.32\linewidth]{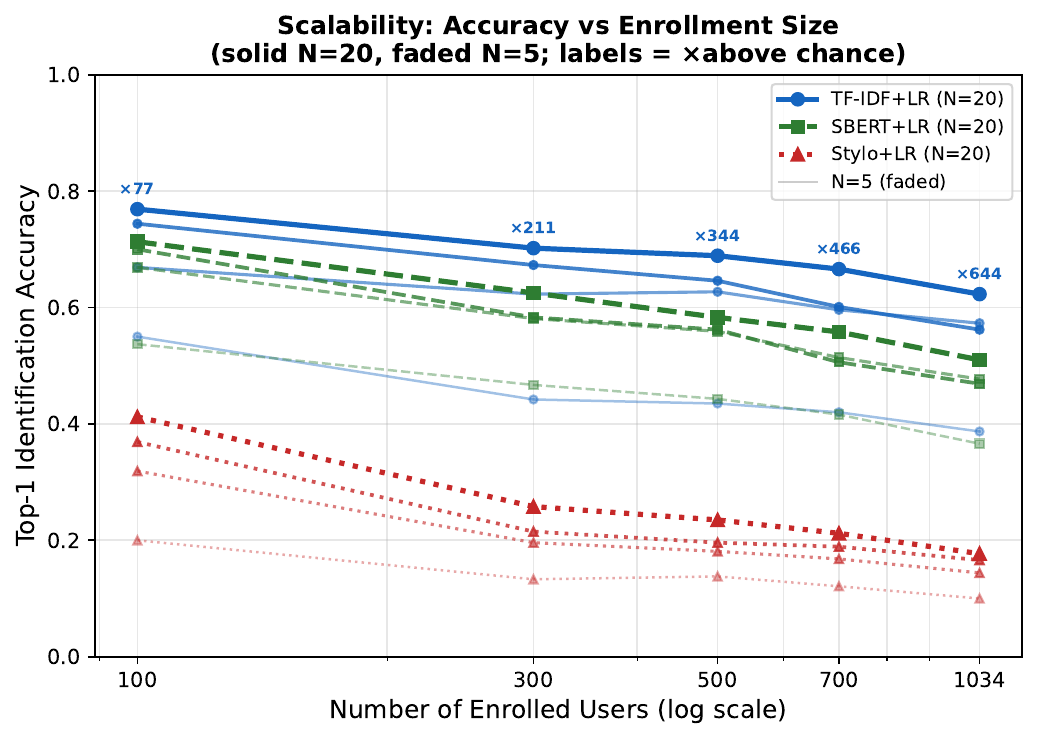}
  \caption{%
    \textit{Left:} Uniqueness-consistency trade-off. Each point represents one method (mean $\pm$ std, 5 folds). Stylo-NN (red) 
    isolates the paradox: highest $d'$ yet lowest Top-1. TF-IDF+LR (orange) achieves optimal verification ($d'$=0.671); Ensemble 
    (blue) achieves optimal identification (Top-1=64.2\%).
    \textit{Middle:} Top-1/3/5 accuracy vs.\ ensemble weight $\alpha$. 
    Plateau between $\alpha{=}0.5$--$0.9$ confirms stability; 
    standardized at $\alpha{=}0.7$.
    \textit{Right:} Top-1 accuracy vs.\ cohort size (log scale). 
    Labels indicate performance factor above chance. TF-IDF consistently outperforms SBERT across all cohort sizes.
  }
  \label{fig:combined}
\end{figure*}

\subsection{The Stylometry Paradox}

The \emph{Stylometry Paradox} refers to the observation that users exhibit distinctive writing styles across a population, yet their style lacks sufficient consistency across contexts for reliable identification. We examine this paradox in the setting of LLM prompts and analyze its implications for prompt-based identity. Stylo-NN produces the highest distributional separation ($d'$=0.950$\pm$0.022) while simultaneously yielding the lowest identification accuracy (Top-1=24.9$\pm$0.9\%). This divergence is resolved through signal detection theory: $d'$ measures the distributional separation between genuine and impostor score populations, whereas Top-1 measures instance-level ranking consistency. 

Users demonstrate \emph{inter-personal distinctiveness} in surface style (high $d'$), yet remain \emph{intra-personally inconsistent}. A user's stylometric profile shifts fluidly with topic, prompt length, and cognitive load~\cite{argamon2007mining}. For example, a user issuing a coding request generates shorter, highly imperative sentences with a higher digit ratio compared to their own factual queries. In aggregate, their distribution remains distinct from other users (high $d'$); however, no single isolated prompt reliably maps back to the user's prototype (low Top-1). Consequently, stylometric features are structurally inadequate as a standalone identification modality for prompts, but provide highly valuable distributional signal when utilized in soft-score verification fusion.

\subsection{Alpha Sweep and Fusion Analysis}
\label{sec:alpha}
Figure~\ref{fig:combined} (middle) illustrates Top-1 accuracy 
as a function of the ensemble weight $\alpha$ (Equation~\ref{eq:ensemble}), evaluated on a held-out fold. At $\alpha{=}0.0$ (pure TF-IDF), Top-1~$=0.621$; at $\alpha{=}1.0$ (pure SBERT), Top-1~$=0.520$. The performance curve peaks at $\alpha{=}0.9$ (Top-1~$=0.646$), 
with our primary ensemble fixed at $\alpha{=}0.7$ 
(Top-1~$=0.642$) a negligible variance confirming robust stability across the range $[0.5, 0.9]$. The extended flat region indicates that SBERT functions as a stable marginal complement to TF-IDF rather than providing an independent discriminative axis. Crucially, while the ensemble improves identification over standalone TF-IDF (64.2\% vs. 63.0\%), it incurs a verification penalty (EER=0.421 vs. 0.381). This stems from the architecture: score-level fusion synthesizes complementary probability rankings (boosting identification), whereas verification relies on the strict geometric separation of the embedding space. TF-IDF naturally achieves this separation: its sparse, high-dimensional token vectors generate distinct genuine and impostor distributions ($d'=0.671$). Integrating SBERT's cosine scores, which possess a weaker geometric boundary ($d'$=0.159), dilutes the spatial separation. Thus, score-level fusion is optimized for identification; maximizing verification requires a jointly trained metric-space model, a direction we reserve for future work (Section~\ref{sec:discussion}).

\begin{table}[t]
\centering
\caption{%
  \textbf{Scalability $\times$ Enrollment.}
  Top-1 accuracy across cohort size and enrollment count $N$.
  \textbf{Bold} $=$ best between TF-IDF and SBERT per cell.
  Chance: $1.0\%$ (100 users) $\to$ $0.097\%$ (1,034 users).
  $N{=}10$ is the stabilisation threshold.
}
\label{tab:scalability}
\setlength{\tabcolsep}{4pt}
\small
\begin{tabular}{|c|l|c|c|c|c|}
\hline
\textbf{Users} & \textbf{Space}
  & \textbf{N=5} & \textbf{N=10} & \textbf{N=15} & \textbf{N=20} \\
\hline
\multirow{3}{*}{100}
  & TF-IDF & \textbf{0.550} & \textbf{0.669} & \textbf{0.744} & \textbf{0.769} \\
\cline{2-6}
  & SBERT  & 0.537 & 0.669 & 0.700 & 0.713 \\
\cline{2-6}
  & \cellcolor{rowgray}Stylo
  & \cellcolor{rowgray}0.200 & \cellcolor{rowgray}0.319
  & \cellcolor{rowgray}0.369 & \cellcolor{rowgray}0.412 \\
\hline
\multirow{3}{*}{300}
  & TF-IDF & 0.442 & \textbf{0.623} & \textbf{0.673} & \textbf{0.702} \\
\cline{2-6}
  & SBERT  & \textbf{0.467} & 0.581 & 0.583 & 0.625 \\
\cline{2-6}
  & \cellcolor{rowgray}Stylo
  & \cellcolor{rowgray}0.133 & \cellcolor{rowgray}0.196
  & \cellcolor{rowgray}0.215 & \cellcolor{rowgray}0.258 \\
\hline
\multirow{3}{*}{500}
  & TF-IDF & \textbf{0.435} & \textbf{0.627} & \textbf{0.646} & \textbf{0.689} \\
\cline{2-6}
  & SBERT  & 0.443 & 0.559 & 0.562 & 0.583 \\
\cline{2-6}
  & \cellcolor{rowgray}Stylo
  & \cellcolor{rowgray}0.138 & \cellcolor{rowgray}0.181
  & \cellcolor{rowgray}0.196 & \cellcolor{rowgray}0.235 \\
\hline
\multirow{3}{*}{700}
  & TF-IDF & \textbf{0.420} & \textbf{0.596} & \textbf{0.601} & \textbf{0.666} \\
\cline{2-6}
  & SBERT  & 0.416 & 0.514 & 0.506 & 0.558 \\
\cline{2-6}
  & \cellcolor{rowgray}Stylo
  & \cellcolor{rowgray}0.121 & \cellcolor{rowgray}0.168
  & \cellcolor{rowgray}0.189 & \cellcolor{rowgray}0.212 \\
\hline
\multirow{3}{*}{1,034}
  & TF-IDF & \textbf{0.387} & \textbf{0.573} & \textbf{0.562} & \textbf{0.623} \\
\cline{2-6}
  & SBERT  & 0.366 & 0.477 & 0.469 & 0.510 \\
\cline{2-6}
  & \cellcolor{rowgray}Stylo
  & \cellcolor{rowgray}0.100 & \cellcolor{rowgray}0.144
  & \cellcolor{rowgray}0.166 & \cellcolor{rowgray}0.177 \\
\hline
\end{tabular}
\end{table}

\subsection{Scalability}
Table~\ref{tab:scalability} and Figure~\ref{fig:combined} (right) report system performance across gallery sizes ranging from 100 to 1,034 users. As expected, Top-1 accuracy decreases monotonically with gallery size across all feature spaces. This degradation is a well-documented phenomenon in both authorship attribution~\cite{luyckx2011effect, tschuggnall2019reduce} and soft biometric identification~\cite{dantcheva2015else}, where inter-class confusability grows as the candidate pool expands and random chance itself falls as $1/N$. Luyckx and Daelemans~\cite{luyckx2011effect} systematically demonstrated this effect for stylometric classifiers, showing steep accuracy declines as author sets scale beyond small laboratory sizes.

Despite a $10.3\times$ increase in gallery size, the absolute Top-1 accuracy drop for TF-IDF is 14.6 percentage points ($76.9\%$ to $62.3\%$ at $N{=}20$), indicating that the lexical signal retains discriminative value at scale. The ensemble outperforms standalone TF-IDF by 2-4 percentage points across all cohort sizes, a margin that remains approximately stable rather than diverging or collapsing; however, this modest gap should be interpreted alongside the verification trade-off discussed in Section~\ref{sec:alpha}.

\subsection{Enrollment Requirements and Feature-Space Crossover}
\label{sec:enrollment}

Table~\ref{tab:scalability} tracks Top-1 accuracy for 
TF-IDF+LR, SBERT+LR (frozen encoder), and Stylo+LR across a full matrix of deployment scales and user enrollment counts. All pipelines utilize LogisticRegression for direct comparative validity.

\noindent\textbf{Enrollment Stabilization.} N=10 represents a universal stabilization threshold. The performance delta from N=5$\to$N=10 remains consistent across all scales and feature spaces: $+$11.9-19.2pp for TF-IDF, $+$11.4-13.8pp for SBERT, and $+$4.6-11.9pp for Stylo. Beyond 10 prompts, marginal gains diminish sharply, establishing it as the practical minimum for deployment.

\noindent\textbf{Representation Crossover.} In small deployments with sparse enrollment ($\leq$300 users, N=5), SBERT outpaces or ties TF-IDF (e.g., 0.467 vs. 0.442 at 300 users, N=5). However, at larger scales ($\geq$300 users, N$\geq$10), TF-IDF fundamentally dominates. This crossover validates the mechanics of the score-level ensemble: SBERT provides critical fallback signals when a user's lexical map is too sparse, while TF-IDF drives accuracy once the vocabulary fingerprint reaches critical mass.

\noindent\textbf{Stylometric Variance.} In direct contradiction to standard assumptions, Stylo+LR requires volume to function; it improves dramatically from 0.200 to 0.412 (100 users) and 0.100 to 0.177 (1,034 users) as N scales from 5 to 20. A single prompt lacks the linguistic surface area to reliably estimate part-of-speech ratios or vocabulary richness.

\subsection{TF-IDF Vocabulary Size Ablation}
\label{sec:vocab_ablation}

To isolate the efficiency of the primary lexical component, we mapped vocabulary size constraints (\texttt{max\_features}) 
across $\{$5K, 10K, 20K, 30K, 50K, 100K$\}$ on a held-out 
fold (Table~\ref{tab:vocab_ablation}). While accuracy scales monotonically, the inflection points dictate deployment feasibility. The most significant leap occurs between 5K$\to$20K (+5.1pp), plateauing slightly at the 20K$\to$30K boundary (+0.4pp). Expanding to 100K yields an additional 3.2pp, but demands a 3.3$\times$ increase in feature dimensionality and 3$\times$ longer training cycles. The 30K parameter establishes an optimal efficiency threshold, capturing the discriminative core of user habits while filtering out high-noise, long-tail tokens.

\begin{table}[t]
\centering
\caption{%
  \textbf{TF-IDF Feature Count Ablation.}
  Effect of \texttt{max\_features} on Top-$k$ accuracy and training time
  (single fold, 1,034 users, $N{=}10$).
  $^*$ = baseline used in primary experiments.
}
\label{tab:vocab_ablation}
\small
\setlength{\tabcolsep}{5pt}
\begin{tabular}{|c|c|c|c|c|}
\hline
\textbf{max\_feat.}
  & \textbf{Top-1 $\uparrow$}
  & \textbf{Top-3 $\uparrow$}
  & \textbf{Top-5 $\uparrow$}
  & \textbf{Time (s)} \\
\hline
5,000   & 0.572 & 0.681 & 0.718 & 7.1  \\
\hline
10,000  & 0.601 & 0.708 & 0.742 & 12.3 \\
\hline
20,000  & 0.623 & 0.723 & 0.756 & 18.4 \\
\hline
\cellcolor{rowgray}30,000$^*$
  & \cellcolor{rowgray}0.627
  & \cellcolor{rowgray}0.730
  & \cellcolor{rowgray}0.765
  & \cellcolor{rowgray}22.3 \\
\hline
50,000  & 0.645 & 0.745 & 0.775 & 33.1 \\
\hline
100,000 & \textbf{0.659} & \textbf{0.757} & \textbf{0.784} & 94.7 \\
\hline

\end{tabular}
\end{table}
\subsection{Adversarial Robustness}

\noindent\textbf{Evaluation Protocol.} 
To rigorously test the stability of the biometric signature, adversarial transformations were applied to held-out test prompts and processed through the full TF-IDF+SBERT ensemble ($\alpha$=0.7). Evaluating the full pipeline ensures the attacker must bypass the complete multi-modal defense rather than an isolated semantic vector.

As illustrated in Figure 4 and detailed in Table~\ref{tab:adversarial}, 
the hierarchical severity of these attacks provides critical mechanistic insight into the identity signal. Synonym substitution (20\%) fails to disrupt the fingerprint 
(Top-1=0.641, $\Delta$=$-$0.001); because core nouns, verbs, and structural phrasing remain intact, TF-IDF absorbs the minor adjective permutations seamlessly. Homoglyph injection (20\%) causes measurable degradation 
(Top-1=0.588, $\Delta{=}{-}0.054$), as the corrupted subword tokenization bypasses SBERT's semantic mapping, though TF-IDF retains partial resilience via adjacent bigram boundaries. 

However, full semantic paraphrase triggers a catastrophic systemic failure. Erasing the original lexical surface drives Top-1 accuracy down to 0.429 ($\Delta$=$-$0.213) and spikes the EER to 0.703. When forced to analyze structurally novel text with an identical semantic payload, the ensemble collapses. This conclusively validates the Lexical Stability Hypothesis: \emph{the biometric fingerprint resides exclusively in the surface syntax, and vanishes entirely when semantically equivalent rephrasing occurs}.

\begin{table*}[t]
\centering
\caption{%
  \textbf{Adversarial Robustness — Full Rate Sweep} (5-fold, mean$\pm$std).
  All attacks applied to Ensemble ($\alpha{=}0.7$).
  $\Delta$Top-1 and $\Delta$EER relative to unattacked baseline (rate $= 0\%$).
  Paraphrase is a full surface rewrite (rate $= 100\%$).
}
\label{tab:adversarial}
\adjustbox{max width=\textwidth}{%
\begin{tabular}{|c|c|c|c|c|c|c|c|c|c|}
\hline
\textbf{Attack} & \textbf{Rate}
  & \textbf{Top-1 $\uparrow$} & \textbf{Top-3 $\uparrow$} & \textbf{Top-5 $\uparrow$}
  & \textbf{AUC $\uparrow$} & \textbf{EER $\downarrow$} & \textbf{d$'$}
  & \textbf{$\Delta$Top-1} & \textbf{$\Delta$EER} \\
\hline
\multicolumn{1}{|l|}{\cellcolor{rowgray}None (baseline)} & \cellcolor{rowgray}—
  & \cellcolor{rowgray}0.642$\pm$0.004 & \cellcolor{rowgray}0.755$\pm$0.004 & \cellcolor{rowgray}0.793$\pm$0.005
  & \cellcolor{rowgray}0.603$\pm$0.005 & \cellcolor{rowgray}0.421$\pm$0.005 & \cellcolor{rowgray}0.471$\pm$0.009
  & \cellcolor{rowgray}— & \cellcolor{rowgray}— \\
\hline
\multirow{4}{*}{Synonym}
  & 5\%
  & 0.641$\pm$0.004 & 0.755$\pm$0.004 & 0.793$\pm$0.005
  & 0.603$\pm$0.005 & 0.421$\pm$0.005 & 0.471$\pm$0.009
  & $-$0.001 & $\phantom{+}$0.000 \\
\cline{2-10}
  & 10\%
  & 0.642$\pm$0.004 & 0.756$\pm$0.004 & 0.793$\pm$0.005
  & 0.603$\pm$0.005 & 0.421$\pm$0.005 & 0.469$\pm$0.009
  & $\phantom{-}$0.000 & $\phantom{+}$0.000 \\
\cline{2-10}
  & 20\%
  & 0.641$\pm$0.003 & 0.755$\pm$0.003 & 0.793$\pm$0.005
  & 0.603$\pm$0.005 & 0.422$\pm$0.006 & 0.465$\pm$0.009
  & $-$0.001 & $+$0.001 \\
\cline{2-10}
  & 30\%
  & 0.641$\pm$0.004 & 0.755$\pm$0.004 & 0.793$\pm$0.005
  & 0.603$\pm$0.005 & 0.422$\pm$0.005 & 0.462$\pm$0.010
  & $-$0.001 & $+$0.001 \\
\hline
\multirow{4}{*}{Homoglyph}
  & 5\%
  & 0.635$\pm$0.001 & 0.752$\pm$0.005 & 0.789$\pm$0.006
  & 0.620$\pm$0.004 & 0.408$\pm$0.003 & 0.490$\pm$0.007
  & $-$0.007 & $-$0.013 \\
\cline{2-10}
  & 10\%
  & 0.622$\pm$0.005 & 0.739$\pm$0.003 & 0.779$\pm$0.006
  & 0.628$\pm$0.004 & 0.403$\pm$0.004 & 0.503$\pm$0.007
  & $-$0.020 & $-$0.018 \\
\cline{2-10}
  & 20\%
  & 0.588$\pm$0.004 & 0.708$\pm$0.008 & 0.750$\pm$0.008
  & 0.621$\pm$0.003 & 0.409$\pm$0.005 & 0.438$\pm$0.007
  & $-$0.054 & $-$0.012 \\
\cline{2-10}
  & 30\%
  & 0.564$\pm$0.004 & 0.680$\pm$0.005 & 0.720$\pm$0.005
  & 0.596$\pm$0.008 & 0.423$\pm$0.009 & 0.313$\pm$0.015
  & $-$0.078 & $+$0.002 \\
\hline
\multicolumn{1}{|l|}{\cellcolor{rowgray}Paraphrase} & \cellcolor{rowgray}100\%
  & \cellcolor{rowgray}0.429$\pm$0.005 & \cellcolor{rowgray}0.553$\pm$0.006 & \cellcolor{rowgray}0.606$\pm$0.006
  & \cellcolor{rowgray}0.236$\pm$0.005 & \cellcolor{rowgray}0.703$\pm$0.003 & \cellcolor{rowgray}0.967$\pm$0.025
  & \cellcolor{rowgray}$-$0.213 & \cellcolor{rowgray}$+$0.282 \\
\hline
\end{tabular}%
}
\end{table*}

\begin{figure}[t]
  \centering
  \includegraphics[width=\columnwidth]{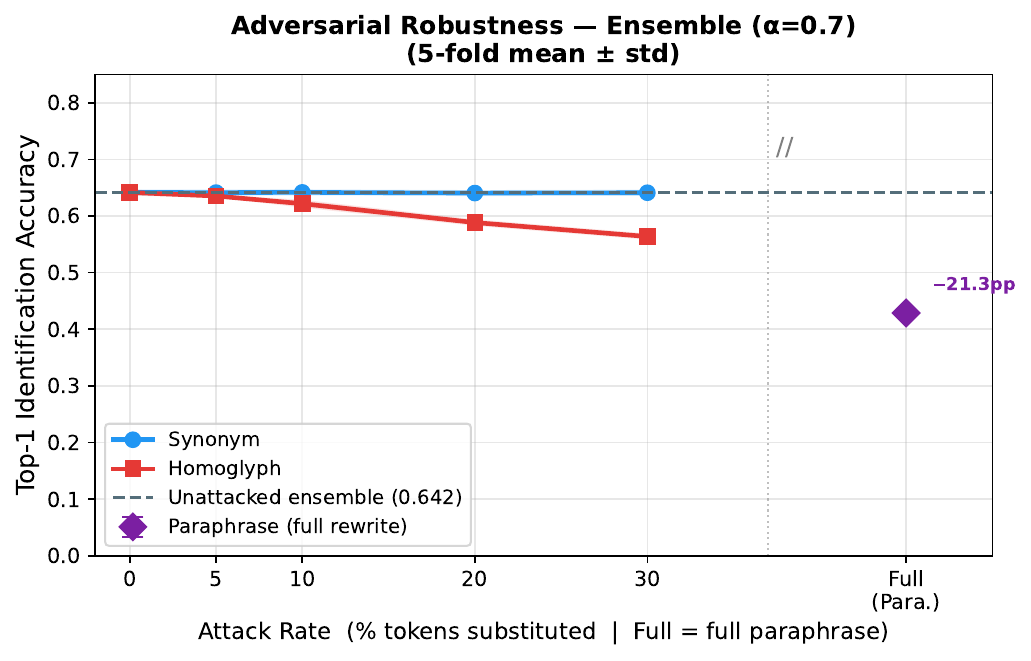}
  \caption{Adversarial robustness modeling. \textit{Left:} Top-1 accuracy vs.\ token substitution rates. 
\textit{Right:} Impact of complete threat vectors against the unattacked Ensemble baseline (0.642). Full paraphrase executes the most severe degradation (Top-1=0.429, EER=0.703), proving the fingerprint is bound to the lexical surface rather than semantic intent.}
  \label{fig:attack}
\end{figure}

%% file: 7_discussion.tex
\section{Discussion}
\label{sec:discussion}

\subsection{Prompt Idiolect}

Prompt idiolect refers to the consistent, user-specific patterns embedded in prompt construction spanning lexical choice, syntactic structure, and interaction style that persist across tasks and can serve as a basis for behavioral identification.  Our findings collectively support a unifying perspective: \emph{prompt identity is fundamentally a lexical-surface biometric}. Specifically, we observe:
\begin{itemize}[noitemsep]
  \item \textbf{Lexically encoded:} TF-IDF ($d'$=0.671) severely outperforms SBERT ($d'$=0.159).
  \item \textbf{Token-surface dependent:} Full semantic paraphrase induces catastrophic degradation (ensemble Top-1=0.429), whereas synonym substitution (20\%) causes negligible impact (Top-1=0.641).
  \item \textbf{Inter-personally distinctive:} Stylometric features yield high distributional separation (Stylo-NN $d'$=0.950).
  \item \textbf{Intra-personally inconsistent:} Stylometric features simultaneously fail at instance-level ranking (Stylo-NN Top-1=24.9\%).
  \item \textbf{Resistant to semantic integration:} Feature-level 
fusion with SBERT ($d'$=0.097) performs worse than 
either modality in isolation.
\end{itemize}

\subsection{Biometric Baseline Calibration}

PromptPrint is a foundational discovery study 
characterizing the existence, location, and adversarial 
robustness of a novel soft biometric signal i.e. prompting behavior. 
An EER of 0.381 (TF-IDF standalone), achieved using only 20 
prompts, no dedicated enrollment procedure, and noisy IP-based 
identity labels, establishes a meaningful preliminary signal for 
a previously unstudied modality. As a passive, enrollment-free 
channel operating on unconstrained in-the-wild text, PromptPrint 
is most appropriately contextualized within the soft biometrics 
paradigm~\cite{dantcheva2015else}, where modalities such as 
gait~\cite{sprager2015inertial} and writing style~\cite{ fridman2016active} 
report EERs in the 0.15-0.35 range under mature, controlled 
conditions. Our baseline of 0.381, achieved without any dedicated 
enrollment protocol and under noisy label conditions, is consistent 
with early-stage discovery results in comparable soft biometric 
modalities. Transitioning this signal to a deployment-ready system 
would necessitate dedicated metric-learning objectives, expanded 
enrollment sets, and adversarial hardening, which we designate as 
critical vectors for future research.

\subsection{Evaluation Limitations}

Score-level fusion successfully optimizes identification but degrades verification relative to standalone TF-IDF (EER=0.421 vs. 0.381). This divergence occurs because the two objectives require distinct topological representations: identification leverages complementary probability distributions, whereas verification demands the strict geometric separation of genuine and impostor embeddings. TF-IDF natively achieves this separation ($d'$=0.671), but score mixing actively dilutes it ($d'_\text{ens}$=0.471). Future architectures could close this gap by applying triplet or contrastive loss directly to sparse, TF-IDF-inspired representations, optimizing both ranking and thresholding within a unified metric space. Recent contrastive learning approaches for disentangling authorship from content ~\cite{huertas2024isolating}, ~\cite{rivera2021learning} suggest promising directions for such joint optimization

Biometric evaluation protocols recommend reporting True Accept Rates (TAR) at fixed operational thresholds. Derived via ROC curve interpolation, our best-performing TF-IDF configuration yields 
TAR=16.6\%$\pm$4.3\% at FAR=1\%, TAR=3.4\%$\pm$0.9\% 
at FAR=0.1\%, and TAR=2.3\%$\pm$1.6\% at FAR=0.01\%.

%% file: 5_Conclusion.tex
\section{Conclusion}
\label{sec:conclusion}

This work formalizes prompt-based identity as a viable behavioral biometric, demonstrating that users leave a persistent, highly stable lexical fingerprint within their instrumental interactions with large language models. Our evaluation confirms that a constrained 30,000-feature lexical representation successfully isolates this identity signal, achieving 64.2\% Top-1 accuracy across a 1,034-user 
cohort outperforming random chance by a factor of 664. 

Through rigorous ablation, scalability testing, and adversarial localization, we conclusively map this identity fingerprint to the token-surface layer rather than the semantic payload. Because full semantic paraphrase effectively sanitizes the behavioral signal (degrading Top-1 accuracy to 0.429 and driving EER to 0.703), it is evident that generative identity resides in the idiosyncrasies of surface phrasing, which semantic rewriting intrinsically destroys. While score-level fusion optimizes closed-set identification, the sparse lexical component structurally dominates open-set verification. 

Ultimately, demonstrating the stability of prompt as behavioral biometric highlights an important duality in the AI ecosystem. On one hand, it can support new forms of low-friction authentication and anomaly detection. On the other hand, it also reveals a potential surveillance risk, as users may be unaware that routine prompts can act as digital identifiers. By defining this threat model and releasing our evaluation protocols, we provide a foundation for future work in two directions: developing practical uses of prompt-based identification and designing robust privacy-preserving countermeasures.